# Clinical Intent Extraction: A FHIR-Aligned Representation and the CIRCA Benchmark

*Alexander Apartsin[a] · Yehudit Aperstein[b,*]*
*[a]Holon Institute of Technology, Holon, Israel [b]Afeka College of Engineering, Tel Aviv, Israel*
*[*]Corresponding author:* apersteiny@afeka.ac.il

**Abstract**

Prospective clinical actions, the follow-ups, orders, referrals, and instructions that determine what happens to a patient next, are annotated today in thin fragments across incompatible corpora: each records a text span and one coarse category. We introduce **Clinical Intent Extraction (CIE)**, the task of recovering these actions as complete structured records, and the **Clinical Intent Representation (CIR)**, which decomposes each action into its verb, type, coded target, timing, and condition, and adds two axes prior datasets do not jointly represent: request-intent, the authority behind the action (proposal, plan, order, or option, aligned to HL7 FHIR), and modality, a seven-valued scale of clinical strength. Re-expressing five heterogeneous corpora (CLIP, MedDec, ap_parsing, PaniniQA, SIMORD) in the CIR yields **CIRCA**: 10,011 harmonized intents spanning two note distributions, with a human-validated subset, source-to-CIR crosswalks, and a deterministic FHIR R4 mapper. CIRCA is built by three-model consensus that auto-accepts high-agreement intents and routes the rest to human review; the audited agreement stratum matches human decisions 88.4% of the time. Benchmarking five existing models without task-specific training exposes the gap CIRCA targets: given the span, they label type well (85 to 91%) but get all four closed fields right only 18 to 35% of the time. All artifacts are released, with MIMIC-derived layers shared as stand-off annotations under PhysioNet credentialed access.

**Keywords:** clinical NLP, information extraction, dataset harmonization, FHIR, large language models, inter-annotator agreement, human-in-the-loop

## 1 Introduction

A discharge summary does two things at once: it records what happened, and it commits the patient to what should happen next. The second, prospective layer ("repeat the CBC (complete blood count) in two weeks," "follow up with cardiology," "hold the metoprolol if the pressure drops") determines whether care continues safely after the patient leaves. Yet these prospective actions are frequently not completed: a large share of laboratory and imaging results go without documented follow-up[1], and roughly 40% of hospital discharges leave test results still pending, many of them actionable and unnoticed by the responsible clinician[2], a gap compounded by well-documented deficits in the discharge handoff[3] and answered by repeated calls for systems that explicitly capture and track actionable follow-up recommendations[4]. Extracting it in structured form is a prerequisite for follow-up tracking, care-gap detection, and interoperable care plans. This prospective layer is increasingly framed as the next computable object in healthcare IT: earlier generations of clinical systems made first the record and then the patient's clinical state machine-readable, and the emerging layer is patient-specific clinical *intent*, the future-directed actions a note commits to[5]. Existing interoperability standards can carry such intent once it is already structured, but they do not recover it from the free-text clinical communication where it originates, which is the gap a clinical intent-extraction task must close.

Several corpora annotate parts of this layer. CLIP[6] labels discharge-note sentences with seven action-item aspects; MedDec[7] marks medical-decision spans; ap_parsing[8] segments assessment-and-plan action items; PaniniQA[9] annotates discharge-instruction content; and SIMORD / MEDIQA-OE[10] extracts structured orders from consult dialogue (order annotations layered on the public ACI-Bench corpus). Each is valuable, and each is an island: the label sets, note types, and granularities all differ, and none shares a notion of how *authoritative* or *strong* an intended action is. To our knowledge, no existing resource lets a downstream system reason about prospective actions uniformly across them.

We argue the missing piece is not another corpus but a *representation* and a *harmonization*. We define CIE and a FHIR-informed CIR, and we build **CIRCA** (Clinical Intent Representation, Cross-corpus Annotations), a dataset that re-expresses the five heterogeneous corpora in it. The two axes are associated but non-redundant, so a single-axis schema would silently discard safety-critical intents such as conditional or prohibited actions (§3.3). Our contributions:

1. **The Clinical Intent Extraction (CIE) task and the Clinical Intent Representation (CIR).** A closed action taxonomy, a dual target (verbatim span, normalized concept, standard code), separated time / frequency / duration, and two axes that prior clinical-note datasets do not represent as separate structured attributes: request-intent (adapted from FHIR RequestIntent) and modality (clinical strength), with deterministic serialization of CIR records into FHIR R4 request resources where representable.
2. **A harmonized multi-corpus dataset.** 10,011 CIR intents over 1,185 notes and 942 patients from five heterogeneous corpora (differing in annotation schema, granularity, and note type) spanning two note distributions, credentialed MIMIC-III text and a public non-MIMIC dialogue corpus, characterized by per-field distributions and per-corpus profiles, with a human-reviewed core of 2,145 validated intents plus 255 reviewed negatives, per-layer licensing and provenance, and source-to-CIR crosswalks. It is built by seeding three LLMs with the existing corpora as anchors, re-extracting and labeling into the CIR, and routing disagreements to span-level human review (§4).
3. **A baseline evaluation of existing large language models.** No training; existing models are evaluated against the human gold on two tracks: span-level labeling (given the span, predict the CIR fields) and whole-note (note-only) extraction (find the spans without anchors).

## 2 Related work

**Clinical action and order extraction.** CLIP, MedDec, ap_parsing, PaniniQA and SIMORD each target a facet of prospective or decision-oriented content, as do the n2c2 medication and temporal corpora[11] and RadGraph for radiology[12]. All are single-corpus and schema-specific. Concretely, these source annotations are thin: each marks a span and assigns it one coarse category, and nothing more. CLIP tags each follow-up span with one of seven follow-up types; MedDec labels each decision span with a single category from the DICTUM decision scheme (eleven types); ap_parsing marks a span as an action item, problem title, or problem description with a coarse action-item class; PaniniQA labels patient-education spans with a jargon category and links them by relation; SIMORD is the richest, adding a free-text reason to each order type. None annotates, as an explicit field, the action verb, the request authority, the directive strength, the timing value, the condition, or a standard code, which are exactly the attributes the CIR makes explicit (Table 1). Noor et al.[13] study the same object, the clinician's prospective follow-up intent, and ground it in spans. They frame it, however, as sentence-level multi-label classification over a flat set of about a dozen institution-specific categories, on a single private clinic corpus. We instead extract each action as a structured record with explicit attributes (action, type, target, timing, condition, request-intent, modality), and we harmonize five existing corpora into one schema. Lau et al.[14] and Steinkamp et al.[15] extract follow-up recommendations from radiology and discharge text. Their task is close to CIE, but they lack the authority and strength axes, cross-corpus harmonization, and a mapping to standards. Table 1 positions CIR against these prior resources across the schema dimensions.

**Prospective intent versus temporal annotation.** Clinical NLP has mature machinery for temporal structure, exemplified by THYME-style annotation[16], but a future time is not the same as an intended action: "the patient will return", "return if symptoms persist", and "a CT was already scheduled" all point forward yet carry different prospective force. CIE makes prospective intent the unit and treats time as one attribute rather than the organizing label, adding whether an action is proposed, planned, ordered, optional, conditional, or prohibited.

**Assertion, certainty, and negation.** The modality axis is related to assertion status (i2b2/n2c2 assertion[17]), negation and context (NegEx[18], ConText[19]), speculation and hedging detection (BioScope[20]); our polarity layer, which maps negated directives to *prohibited*, performs NegEx-style scope detection. The distinction is temporal and pragmatic. Assertion status qualifies a *present finding* (present, absent, hypothetical, possible). Modality here grades the *strength of a prospective directive* (recommended, considered, conditional, optional, prohibited). The two answer different questions about a span and are annotated on different objects.

**Request authority and speech acts.** The request-intent axis concerns the directive status of an action, proposal versus plan versus order versus option, which is closer to speech-act semantics[21] and to the FHIR request pattern[22] than to assertion classification. FHIR represents request authority downstream in structured resources; to our knowledge no clinical-NLP corpus annotates an equivalent authority distinction over free narrative text, which is one of the axes CIR contributes.

**LLMs for annotation across domains.** Large language models have moved from objects of annotation to instruments of it: on many text-classification tasks they rival or surpass crowd workers at a fraction of the cost[23, 24], and prompting strategies such as explain-then-annotate push zero-shot labels toward crowd quality[25]. The shift spans domains, from computational social science[26] and political-text coding[27] to clinical and biomedical information extraction[28, 29, 30], where LLM-drafted labels are paired with expert review. LLMs also construct datasets: in the reminiscence-narrative domain, IRC-Bench is built by an entirely LLM-driven pipeline in which one model performs entity extraction and linking, summary generation, and entity-elided rewriting while a second model validates the outputs[31]. A parallel line treats models as evaluators of their own and each other's outputs: LLM-as-judge protocols approximate human preferences yet inherit position, verbosity, and self-preference biases[32], while self-consistency and self-refinement use inter-sample agreement and self-feedback as internal correctness signals[33, 34], and ensemble self-evaluation ranks models by cross-model consensus without labeled data[35]. The consistent caution is that model confidence is not ground truth: label quality varies sharply by task, so unvalidated generative annotation is unreliable[36], and in healthcare the stakes are compounded by privacy, the need for expertise, and fluently hallucinated labels. CIRCA is built for this gap: it treats agreement among three independent LLMs as a self-evaluation signal for candidate generation, not gold, and routes every disagreement to human validation (§7.1), so mutual consensus accelerates but never substitutes for expert adjudication.

**Harmonization and standards.** Existing harmonization works at the terminology or common-data-model level (mapping entities to OMOP, a common data model for observational health records[37], or to standard code systems for labs (LOINC), clinical concepts (SNOMED CT), and drugs (RxNorm)), or assembles evaluation suites in which each dataset keeps its native schema. Pipelines such as NLP2FHIR convert extracted data into FHIR resources but are engineering systems, not annotated intent datasets. Closest in spirit, Hong et al.[38] standardize heterogeneous clinical annotation corpora to HL7 FHIR for reuse and integration. That line of work aligns entity-level annotations for interoperability. We harmonize *prospective-action* annotations under a request-intent and modality schema, and we release them as a labeled dataset with source-to-CIR crosswalks. FHIR itself[22] defines the RequestIntent value set and the ServiceRequest / CarePlan / RequestGroup resources our schema maps to. Among the resources reviewed here, none annotates clinical notes directly into these axes at scale. We are not aware of a prior resource that re-annotates several clinical corpora into a single prospective-intent schema carrying both a request-authority and a clinical-strength axis.

**Clinical corpora, transfer, and de-identification.** CIRCA is built over MIMIC-III[39] discharge text and the public ACI-Bench[40] dialogue corpus, two distributions that let the harmonized schema be tested for transfer rather than fit to one source. Cross-corpus generalization is a known difficulty in clinical NLP, where domain- and task-adaptive pretraining[41] is a standard response; CIRCA measures transfer at the task and schema level across independently constructed corpora (§7.5). Because MIMIC ships with de-identification placeholders, the pipeline fills them with realistic surrogates, following the identifier-resynthesis line of work[42], so the text stays readable and its date intervals usable without reintroducing patient identity (§8.2).

| Resource | What it natively annotates | prospective intent | action, timing, condition | authority + strength | codes + FHIR | cross-corpus |
|---|---|---|---|---|---|---|
| CLIP | span + 1 of 7 follow-up categories | ● | ○ | ○ | ○ | ○ |
| MedDec | span + 1 of 11 DICTUM decision categories | ◐ | ○ | ○ | ○ | ○ |
| ap_parsing | span role + coarse action-item class | ● | ○ | ○ | ○ | ○ |
| PaniniQA | patient-ed span + jargon category + relations | ● | ○ | ○ | ○ | ○ |
| SIMORD / ACI-Bench | order type + free-text reason | ● | ○ | ○ | ○ | ○ |
| n2c2 med / temporal | medication/time entities + relations | ○ | ◐ | ○ | ◐ | ○ |
| RadGraph | entity + relation graph (radiology) | ○ | ○ | ○ | ◐ | ○ |
| Lau et al. | follow-up recommendation spans | ● | ◐ | ○ | ○ | ○ |
| Steinkamp et al. | follow-up recommendation + timeframe | ● | ◐ | ○ | ○ | ○ |
| **CIR (ours)** | **full structured intent record (9 content fields, 2 novel axes, grounded provenance)** | ● | ● | ● | ● | ● |

***Table 1.*** *What each resource natively annotates, and which CIR attribute groups it captures as explicit labeled fields (● all, ◐ some, ○ none). The groups bundle the CIR fields: action, timing, condition (the structured action detail); authority + strength (the two novel axes, request-intent and modality); and codes + FHIR (standardized terminology and a schema-valid FHIR mapping). The five source corpora target prospective intent but annotate only a span and a single coarse category; among the resources surveyed, CIR is the only one that captures every group and harmonizes across corpora.*

**Synthetic and surrogate clinical data.** Privacy constraints make real clinical text hard to share, motivating synthetic and surrogate data. Neural language models were trained on de-identified notes to emit shareable synthetic records[43], and artificial mental-health and ICU discharge summaries were generated and evaluated for clinical validity and downstream utility[44]. LLM-based methods show that generated labeled data can aid clinical text mining under privacy constraints[45] and that a clinical LLM can be trained largely on synthetic notes while remaining publicly shareable[46]. Closest to our follow-up target, a synthetic outpatient corpus supports reliable action-and-date extraction when learned tagging is separated from deterministic date arithmetic[47]. CIRCA differs in what it releases: rather than distributing model-generated notes, it publishes only labels and cross-corpus mappings, layered onto surrogate-filled MIMIC text processed under a Business Associate Agreement with a zero-retention pipeline (§8).

## 3 The CIE task and the CIR schema

Given a clinical note, CIE extracts every *prospective* clinical intent: an action the note says should happen after the time of writing. Statements about what was already done, and diagnoses about what is true, are out of scope. Each intent is a structured record (Table 2):

| Field | Type | Description |
|---|---|---|
| action | closed (13) | **Required.** obtain, repeat, refer, follow_up, monitor, adjust, start, stop, continue, perform, instruct, schedule, other |
| type | closed (10) | **Required.** appointment_followup, laboratory, imaging, procedure, referral, medication, monitoring, therapy, patient_instruction, other |
| target | dual | **Required.** verbatim span + normalized category + standard code (LOINC / SNOMED / RxNorm); target.code is nullable and is null when no standard code applies to the target |
| request_intent | closed | **Required.** proposal · plan · order · option, a CIR adaptation of the FHIR RequestIntent value set: the *authority* of the request |
| modality | closed | **Required.** the *strength* of the directive; fine values (ordered · planned · recommended · considered · conditional · optional · prohibited) are compared on coarse classes (active / weak / conditional / prohibited) |
| provenance | span(s) | **Required.** verbatim source text grounding every field |
| time / frequency / duration | normalized | **Optional.** WHEN (offset/date), HOW OFTEN (dosing cadence), HOW LONG, kept separate |
| condition | structured | **Optional.** applicability / start / stop trigger ("if the lesion persists") |
| reason | text / coded | **Optional.** the clinical rationale for the intent, when the note states it |
| goal | text | **Optional.** the intended outcome of the intent, when the note states it |

***Table 2.*** *The Clinical Intent Representation. Five fields (action, type, target, request-intent, modality), together with provenance, are required; the rest are populated only when the note states them. The two closed fields request-intent (FHIR authority) and modality (clinical strength) are the schema's two novel axes (§3.3). The _polarity signal used by the normalization layer (§4) is an internal attribute, not a released schema field, and does not appear in released records.*

Separating request-intent from modality matters because a note's *commitment* and its *clinical force* are different signals. "We will follow up in three months" (plan, recommended) and "consider follow-up if symptoms persist" (proposal, conditional) differ on both axes. A downstream care system needs both to prioritize. Two fine modality values, ordered and planned, carry an authority connotation that tends to track request-intent. To keep the strength axis separate from the authority axis, agreement and release compare modality on the coarse strength classes (active / weak / conditional / prohibited); the fine value stays in the record.

### 3.1 Axis value definitions

Because the two axes are the schema's novelty, each value is defined with a typical trigger phrase, so a reader can annotate a sentence directly (Table 3).

| Axis: value | Meaning | Trigger phrasing |
|---|---|---|
| request_intent: proposal | raised for consideration, lowest authority | "we could consider…", "one option would be…" |
| request_intent: plan | the team commits to it, not yet an executable order | "we will follow up…", "plan to repeat…" |
| request_intent: order | an actionable directive issued now | "start warfarin 5 mg", "repeat CBC in two weeks" |
| request_intent: option | an explicitly offered alternative among choices | "may either… or…", "alternatively…" |
| modality: ordered | an active directive, in force now | "take 800 mg every 6 hours" |
| modality: planned | a committed future action | "will schedule a follow-up" |
| modality: recommended | advised but not mandated | "we recommend…", "should…" |
| modality: considered | under consideration, not yet decided | "consider…", "may consider…" |
| modality: conditional | active only when a stated trigger is met | "if INR > 3.5, hold…", "should symptoms persist…" |
| modality: optional | left to patient or clinician discretion | "as needed", "may take… if…" |
| modality: prohibited | the note directs that the action not be done | "avoid…", "do not…", "no driving" |

***Table 3.*** *The two novel axes, value by value, with a definition and a typical trigger phrase for each. Request-intent grades authority (how binding the request is); modality grades clinical strength (how strongly the action is directed). For agreement and evaluation the seven modality values are compared on four coarse classes: active (ordered, planned, recommended), weak (considered, optional), conditional, and prohibited.*

### 3.2 Alignment with FHIR

Because the schema borrows FHIR's request-intent axis, a CIR record maps directly to a FHIR R4 request resource (Table 4). We release a deterministic mapper and validate its output against the R4 schema. Every mappable intent in the dataset (all types except other) produces a **schema-valid** resource; the worked examples in §5.5 show the FHIR resource each intent maps to (Table 7). This is a schema-valid serialization, not a claim of full profile conformance or terminology-binding validation, which are future work.

| CIR field | FHIR element |
|---|---|
| type = medication | MedicationRequest |
| type = laboratory / imaging / procedure / monitoring / therapy / referral / appointment_followup | ServiceRequest |
| type = patient_instruction | CommunicationRequest |
| request_intent (proposal / plan / order / option) | .intent (direct) |
| target.code / target.text | .code / .medicationCodeableConcept |
| time (day offset) | .occurrenceTiming |
| frequency | dosageInstruction.timing |
| modality = prohibited | doNotPerform = true |
| reason, condition | .reasonCode |

*Table 4.* How each CIR field maps to a FHIR R4 request resource. Request-intent is a CIR adaptation of the FHIR RequestIntent value set; the other fields fill standard elements. Reason maps to reasonCode; condition (an applicability or start/stop trigger) is carried as a trigger and is not equivalent to a clinical reason, so its full semantics may need a profile or extension.

### 3.3 Authority and strength as separate axes

Request-intent and modality capture different distinctions, and neither is recoverable from the other. The structural evidence is direct (Table 5): the conditional, prohibited, optional, and considered strengths (986 intents) have no request-intent equivalent, because the FHIR RequestIntent value set cannot express conditionality or prohibition. Collapsing the axes would discard exactly the plan+conditional, order+optional, plan+prohibited, and proposal+conditional cases a follow-up tracker must handle correctly. A second line of evidence counts off-diagonal intents under an authority-to-strength mapping (order and plan map to active strength; proposal and option map to weak strength): across all 10,011 intents, 9% (903) carry a strength the mapping does not predict, and these concentrate in the safety-critical tail. The two axes are strongly associated (Cramér's V ≈ 0.67) while remaining distinct: association measures how far the joint distribution departs from independence, not whether either axis can be recovered from the other.

| request_intent \ modality | recommended | ordered | planned | considered | conditional | optional | prohibited |
|---|---|---|---|---|---|---|---|
| proposal | 24 | 0 | 0 | 96 | 16 | 0 | 0 |
| plan | 1480 | 168 | 2001 | 12 | 139 | 51 | 21 |
| order | 377 | 4916 | 26 | 0 | 98 | 366 | 130 |
| option | 29 | 0 | 1 | 14 | 13 | 27 | 3 |

***Table 5.*** *Joint distribution of request-intent (authority) and modality (strength) over the 10,008 intents with both axes populated (count per cell); three of the 10,011 intents carry a null axis value and are omitted from the crosstab. There is no canonical diagonal between a four-valued authority axis and a seven-valued strength axis. The conditional, prohibited, optional, and considered strengths (986 intents, 9.8%; distinct from the 903 off-diagonal intents discussed in §3.3) have no request-intent equivalent; §3.3 interprets the non-redundancy shown here.*

## 4 Dataset construction

Annotating every prospective intent and all of its attributes across a whole note is slow and heavy: a reader must hold the entire note in mind, decide what is prospective, and fill many fields for each action at once. The method here decomposes that work into short span-level checks. Existing corpora supply partial span annotations to anchor the search, three models re-extract and label into the CIR, model agreement routes effort, and a human confirms or corrects one highlighted span at a time (Figure 1).

**Anchoring.** Each source corpus provides partial span-level annotations: a slice of the intents in a note, expressed in its own schema. These are passed to the models as evidence, not as ground truth. A crosswalk proposes a candidate CIR type per source label, but no source span is assumed correct or in scope.

**LLM span extraction and labeling.** Seeded by the anchors, three independent model families, Claude Sonnet 4.5 (Anthropic), Llama 3.3 70B Instruct (Meta), and Nova Pro (Amazon), served through AWS Bedrock on a BAA-covered zero-retention endpoint, each read the note and produce their own list of CIR intents. They perform genuine re-extraction rather than transcribing the anchors: this is filter and extend, not copying. Where a source over-annotates retrospective content (MedDec and ap_parsing carry many "s/p" (status post, already performed) and "given" spans) the models drop it as non-prospective, and where a source is sparse (PaniniQA, SIMORD) the models add the prospective actions it missed, such as every discharge medication. Deterministic normalization layers then canonicalize time, polarity, medication verbs, and terminology codes so that later comparison reflects genuine disagreement rather than formatting (Appendix A.4).

**Consensus and agreement routing.** Comparing the *i*-th intent of one model to the *i*-th of another is meaningless when models extract different counts, so intents are clustered across models by a weighted similarity of span overlap, normalized target, and type (weights 0.5 / 0.3 / 0.2, accepted above 0.45; the same rule scores the benchmark in §6), and field agreement is measured within clusters. Where all three agree on the critical fields (action, type, target identity, request-intent, modality) and every field is grounded, the label is auto-accepted into the high-confidence silver stratum. Where they disagree, or only some of the three found the intent, the intent is routed to human review.

**Span-level human validation.** A single annotator, a computer-science student trained on the CIR annotation guidelines, confirms or corrects each routed span's CIR label. The annotator sees the highlighted span plus a local context window (about 160 characters on each side) with the model's pre-filled label, not the entire note, so each decision is a short check rather than a full re-reading.

**Annotation was anchor-hinted.** During annotation the models received the source anchors as evidence: for an anchored intent, a model saw the span and a coarse source label, then produced all CIR fields (it inferred the fine fields, including the action verb, the two axes, timing, and condition). Intents the models added beyond the anchors carried no such hint; the model located those spans itself. Two consequences follow for evaluation. The span-level benchmark track gives a model the span and asks for the fields, which matches the condition under which the gold was produced. The whole-note track gives only the note text, so it is strictly harder than the annotation condition: the model must recover the spans it was previously hinted on. The recall gap between the two tracks therefore reflects the combined cost of removing both localization and the source hint.

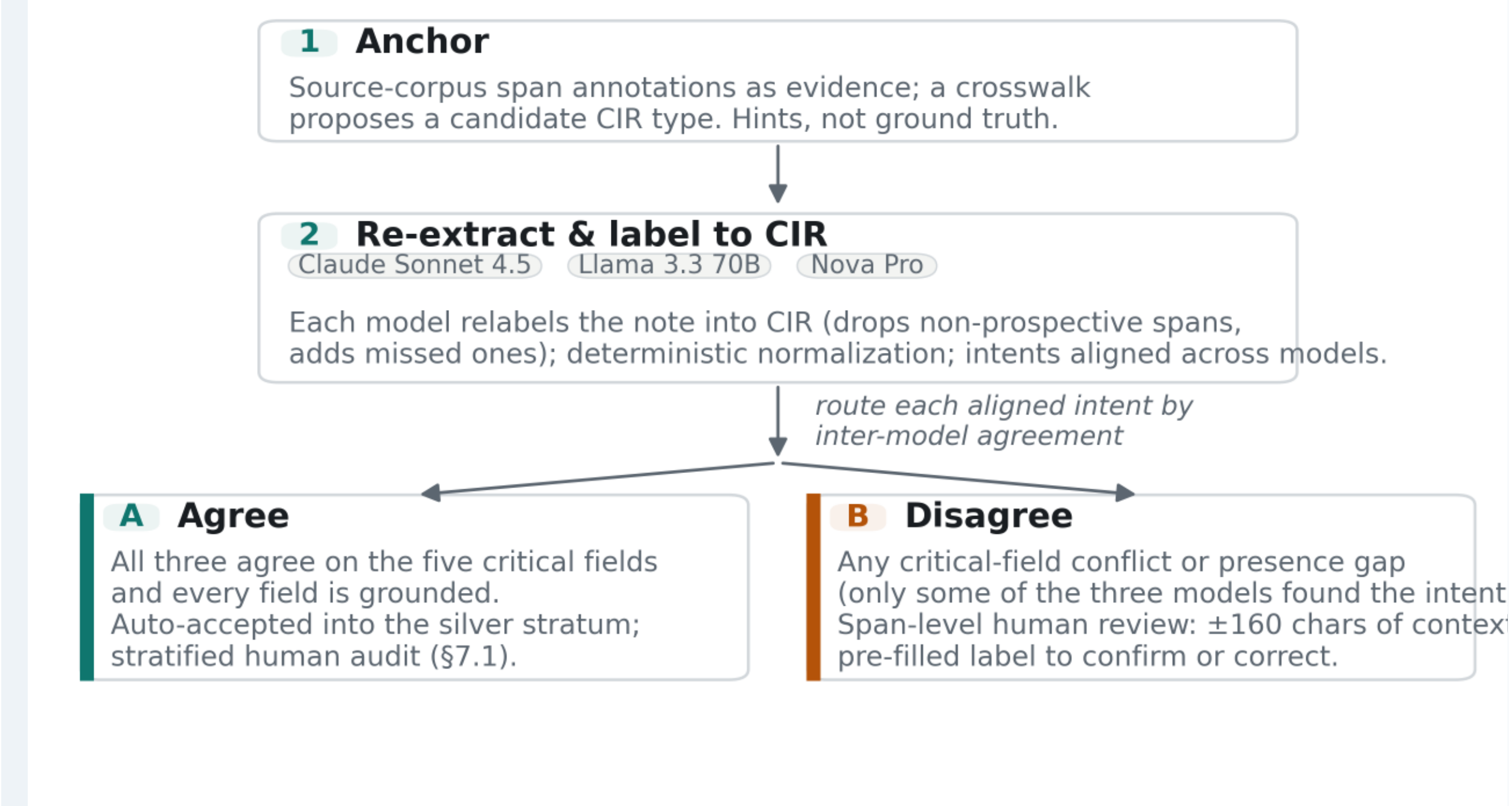

**Figure 1.** Construction pipeline: anchor → re-extract and label to CIR → route by inter-model agreement. Agreement cases (A) are auto-accepted into the silver stratum and audited by stratified sampling (§7.1); disagreement or presence-gap cases (B) go to span-level human review.

The release preserves the provenance tier in every record (§5.4). Because high model agreement is not the same as correctness, the agreement stratum is not treated as gold: a stratified human audit estimates its accuracy directly, and the stratum is released labeled with that measured accuracy rather than as gold (§7.1).

Three of the five models later evaluated as baselines, Claude Sonnet 4.5, Llama 3.3 70B, and Nova Pro, also served in construction (§7.2). We therefore treat those three as construction-involved references rather than independent tests, and read the two gpt-oss models as the independent baselines.

## 5 Dataset composition

CIRCA's harmonized output is 10,011 intents over 1,185 notes across the five corpora (Table 6). The sources are heterogeneous, not a single re-annotation: the four MIMIC-III layers differ in their original annotation task (sentence-level action items, medical-decision spans, assessment-and-plan items, and discharge-instruction QA), and the fifth, ACI-Bench, is public simulated doctor-patient dialogue, contributing a second note distribution and the one layer redistributable with its text. The distributions below expose the class imbalance downstream models must handle.

| Corpus | note type | notes | patients | intents | validated gold | auto-accept / routed | license / access |
|---|---|---|---|---|---|---|---|
| CLIP | MIMIC-III discharge | 250 | 248 | 1,820 | 1,261 | 501 / 1,319 | PhysioNet credentialed |
| MedDec | MIMIC-III discharge | 242 | 216 | 1,596 | 121 | 614 / 982 | PhysioNet credentialed |
| ap_parsing | MIMIC-III assessment & plan | 250 | 249 | 3,183 | 561 | 626 / 2,557 | PhysioNet credentialed |
| PaniniQA | MIMIC-III instructions | 250 | 250 | 1,969 | 123 | 503 / 1,466 | PhysioNet credentialed |
| SIMORD / ACI-Bench | ACI-Bench dialogue (public) | 193 | none (simulated) | 1,443 | 79 | 369 / 1,074 | CDLA-Permissive-2.0 (public) |
| **Total** | 2 distributions | 1,185 | 942 MIMIC | 10,011 | 2,145 | 2,613 / 7,398 | mixed |

***Table 6.*** *What the dataset contains, per corpus: notes sampled, distinct patients (MIMIC corpora; ACI-Bench is simulated with no patients), harmonized CIR intents, the human-validated gold intents in the released core, and how the consensus routed each corpus's intents (auto-accepted on three-model agreement / routed to human adjudication). A further 255 span candidates were rejected as not prospective (§5.4). The five corpora draw on 2,270 distinct source notes with negligible cross-corpus overlap, so harmonization is per-corpus re-annotation into one schema rather than multi-annotator overlap; the note/patient overlap and patient-level de-duplication are detailed in §6.*

### 5.1 Field presence

Five fields are mandatory and populated in every intent (action, type, target, request-intent, modality, each with grounding provenance). The remaining fields are filled only when the annotation records a value, and are sparse (Figure 2):

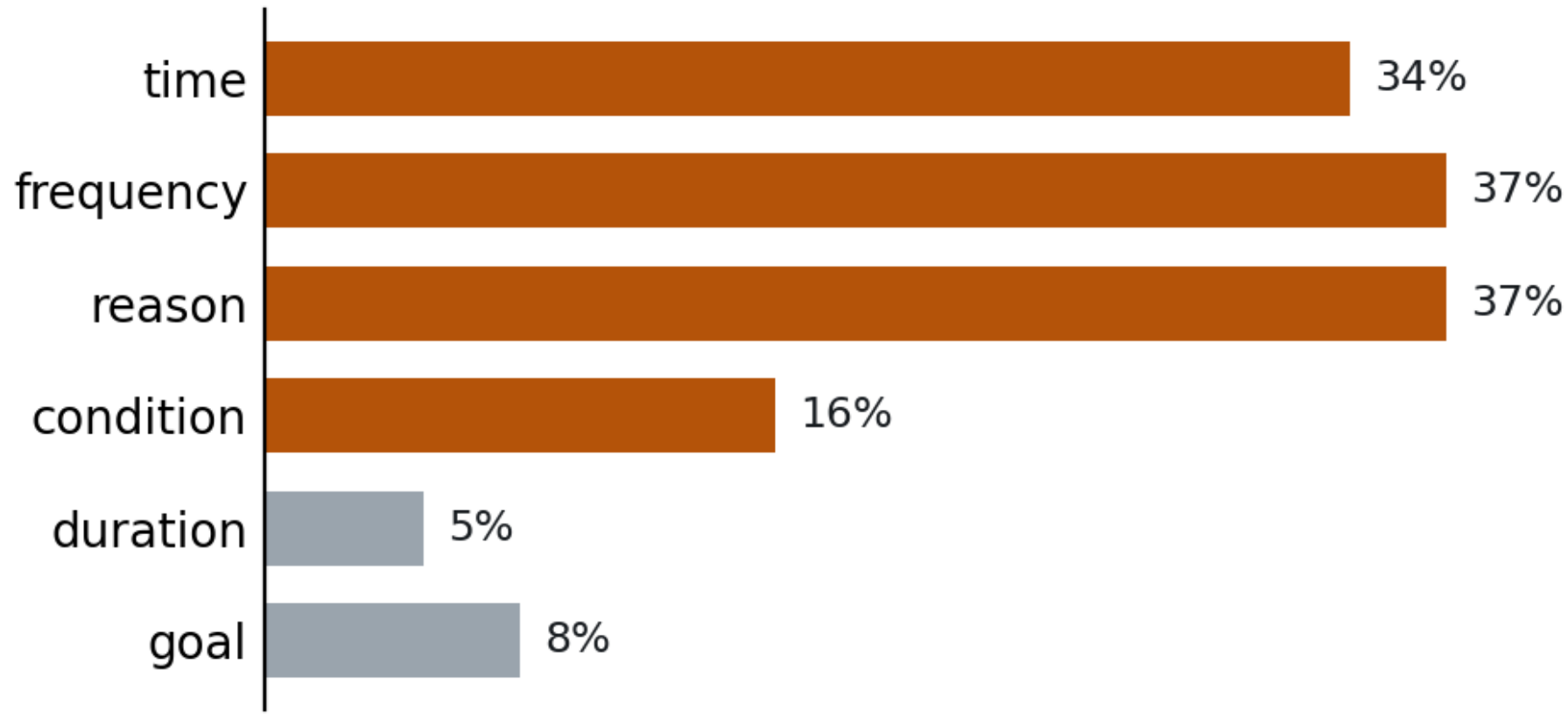

***Figure 2.*** *How often each field is filled across 10,011 intents over 1,185 notes: 34% carry a time and 16% a condition. Fill-rate is a proxy for genuine presence, not a measurement of it; the exhaustive from-scratch pass of §7.6 estimates true presence at 29% for timing and 21% for condition. Timing and conditionality are the sparse, high-value fields where that distinction matters most.*

## 5.2 Value distributions

The four closed fields are heavily skewed toward the common cases, which makes per-class evaluation matter more than raw accuracy (Figure 3).

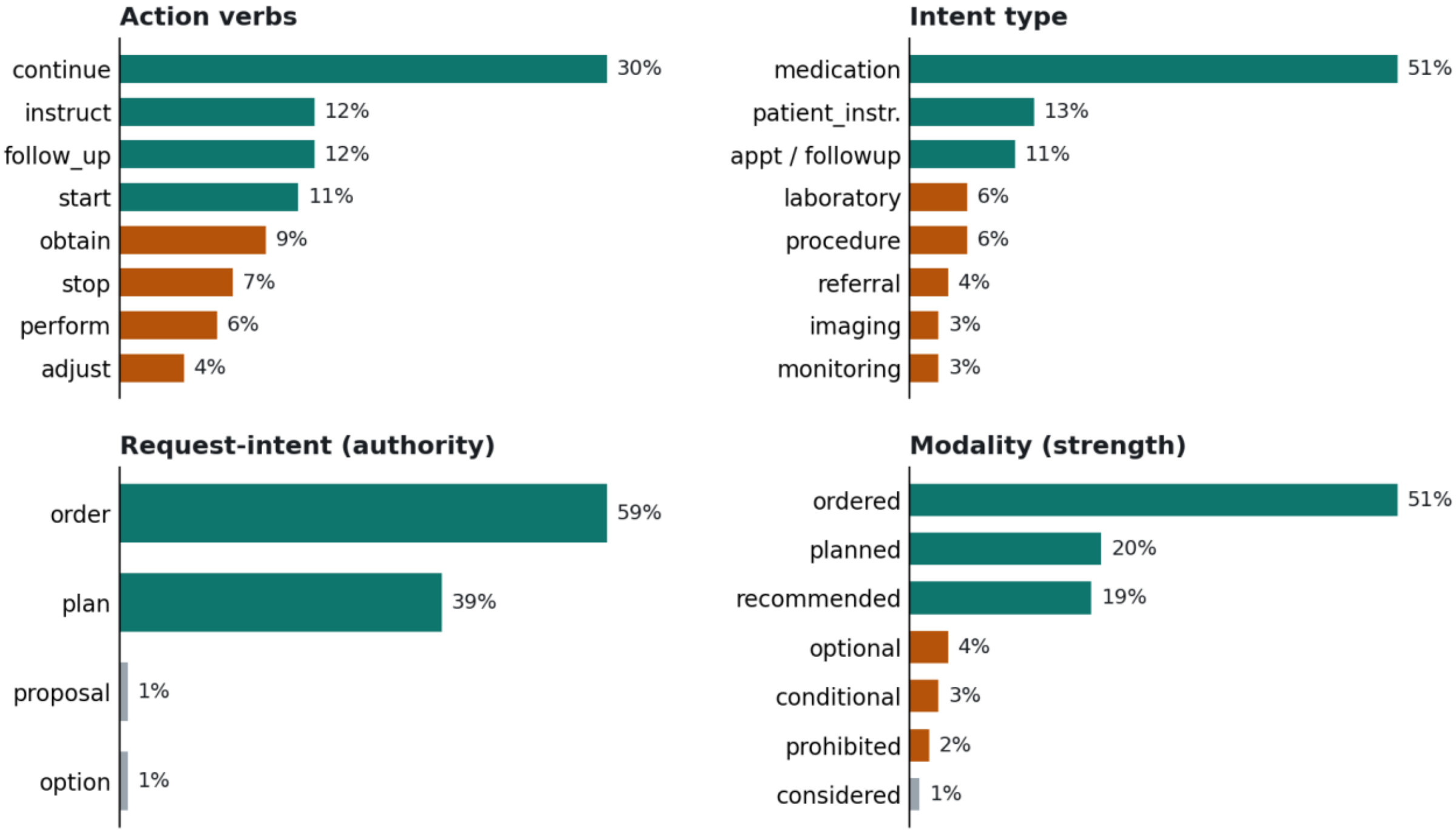

***Figure 3.*** *Value distributions (the most frequent values per panel are shown). Three things stand out. (a) Medication dominates: medication is 51% of all intents, driven by discharge medication lists, and continue alone is 30% of all verbs. (b) Notes commit rather than propose: 98% of intents are plan or order and 90% are recommended, ordered, or planned, so the rarer proposal, conditional, and prohibited cases are the hard, high-value minority. (c) Timing is sparse: most annotations carry no time, condition, or duration value. Given this imbalance, macro-averaged, per-class evaluation matters more than raw accuracy.*

### 5.3 Distributions across the two note sources

The five corpora sample the intent space differently, and the sharpest split is between the two note distributions. Medication dominance is a discharge-summary property: medication is 60% of MedDec and PaniniQA intents, 57% of CLIP, and 46% of ap_parsing, but only 32% of SIMORD, whose ACI-Bench dialogue-derived orders spread across appointment, procedure, laboratory, and imaging instead. This spread makes the cross-corpus transfer protocol (§6) meaningful: a model trained on discharge text is tested on dialogue-derived notes with a genuinely different intent mix.

### 5.4 Human-validated core subset

Every released record carries a filterable provenance tier: **human-adjudicated gold** (a routed span a human confirmed or corrected), **audited high-confidence silver** (an auto-accepted agreement intent whose stratum was human-audited), or **unaudited silver** (released without human inspection). Across the 10,011 intents the consensus auto-accepts 2,613 (26%) on three-model agreement and routes 7,398 (74%) to human adjudication. The release carries a stratified human-reviewed core of 2,400 candidate spans (450 from the agreement stratum, 1,651 from the disagreement stratum, 299 targeting rare modality values); 2,145 are confirmed validated intents and 255 are reviewed non-intent negatives, useful as hard negatives for prospective-versus-retrospective discrimination. Section 4 describes how the tiers are produced; §7.1 reports the audit validating the silver.

### 5.5 Worked examples

Each worked example (Table 7) shows the source sentence, the CIR record it produces, the FHIR resource it maps to (§3.2), and what a downstream care system can do with it. The examples illustrate the representation, not validated executable behavior: a schema-valid resource is a serialization, not a clinical safety rule.

| action | type | target (code) | timing | condition | reason | req. intent | modality |
|---|---|---|---|---|---|---|---|
| **Ex. 1** (ACI-Bench): *"If she is not better at that time, we will obtain an MRI."* | | | | | | | |
| obtain | imaging | MRI (LOINC 24725-4) | – | "if she is not better at that time" | – | plan | conditional |
| → **ServiceRequest**. **Use:** hold the MRI until the condition is met. | | | | | | | |
| **Ex. 2** (ACI-Bench): *"She will take Motrin 800 mg, every 6 hours with food, for two weeks."* | | | | | | | |
| start | medication | Motrin 800 mg → ibuprofen (RxNorm 5640) | q6h; P2W | – | – | order | ordered |
| → **MedicationRequest**. **Use:** reconciliation and adherence; "still active in two weeks?" becomes computable. | | | | | | | |
| **Ex. 3** (synthetic): *"Start warfarin 5 mg daily for atrial fibrillation; hold for INR above 3.5."* | | | | | | | |
| start | medication | warfarin (RxNorm 11289) | daily | "hold for INR above 3.5" (stop) | atrial fibrillation | order | ordered |
| → **MedicationRequest**. **Use:** an INR-monitoring alert consumes the stop-condition. | | | | | | | |
| **Ex. 4** (synthetic): *"No driving for one month. Avoid alcohol while taking this medication."* (one sentence, two intents) | | | | | | | |
| instruct | patient_instruction | driving restriction | P1M | – | – | order | prohibited |
| instruct | patient_instruction | alcohol use | – | "while taking this medication" | – | order | prohibited |

| action | type | target (code) | timing | condition | reason | req. intent | modality |
|---|---|---|---|---|---|---|---|
| → two **CommunicationRequest**. **Use:** suppress "did you drive?" prompts for a month; flag alcohol use while the medication is active. | | | | | | | |

***Table 7.*** *Worked examples: source sentence to CIR record to FHIR resource, with the downstream behavior each enables. A dash (–) marks a field the intent does not populate, which makes the schema's sparsity visible. Example 4 is a paired prohibition that yields two intents (4a, 4b) from one sentence. Examples tagged ACI-Bench are verbatim from the public, simulated corpus[40]; those tagged synthetic are constructed in the style of the credentialed MIMIC corpora, which the data-use agreement does not permit us to reproduce.*

## 6 The CIE benchmark

CIRCA doubles as a reproducible benchmark for CIE: a fixed task, two evaluation tracks, patient-level splits, a matching rule, and a reference evaluate.py with dataset loaders. The task is intent extraction: the input is a note and the output is a set of CIR records.

**Tracks and splits.** The **span-level** track gives a model each gold span with its local context and asks for the four closed fields, so coverage is 100% and there is no extraction step. The **whole-note** track gives only the note text; the model must find the spans itself, which is strictly harder than the anchored condition under which the gold was produced (§4). Splits are patient-level, never note-level, to prevent leakage across a patient's notes. Of the 470 human-validated notes (2,145 gold intents), 20 notes (206 intents) form a prompt-tuning dev set and 450 notes (1,939 intents) the test split; the span-level track draws 600 spans from the validated core. A **cross-corpus transfer** protocol (train on four corpora, test on the held-out fifth) is released for future work; this paper's main benchmark evaluates existing models without training, and §7.5 adds a light trained cross-corpus baseline as a diagnostic.

**Matching rule.** A predicted intent matches a reference by a weighted similarity, 0.5 times span-token overlap (token Jaccard) plus 0.3 times target agreement plus 0.2 times type agreement, accepted above a threshold of 0.45, with a code cannot-link that blocks any match whose resolved standard codes differ, and matching is one-to-one. Type contributes to the score but is not required for a match, so type accuracy is measured on matched intents rather than fixed at 100%. The same weighted similarity clusters model outputs during construction (§4).

**Metrics.** The reported score is per-field accuracy on matched intents and the all-four-exact rate (span-level track), and recall with per-field accuracy on matched intents (whole-note track). Precision on the whole-note track is a lower bound, because the gold is sampled per note rather than exhaustively; §7.6 supplies true precision and F1 on an exhaustive 24-note gold. Per-field accuracy is micro-averaged; §7.3 breaks it down by class and by corpus, where the class imbalance (Figure 3) matters most.

To keep the cross-corpus track a valid transfer test, note and patient overlap across the five source corpora was computed: only 13 MIMIC notes are shared across corpora (CLIP∩PaniniQA 9, CLIP∩MedDec 2, MedDec∩PaniniQA 2), 21 patients appear in more than one corpus, and SIMORD/ACI-Bench shares no notes or patients with the MIMIC corpora. Held-out-corpus splits are therefore de-duplicated at the patient level, so neither a note nor a patient leaks between the train and held-out folds.

## 7 Results

### 7.1 Human validation

Span-level human validation confirms that model agreement predicts correctness and that the rater judges rather than rubber-stamps. On the agreement stratum (n=450 audited), the LLM pre-labels matched the human decision 88.4% of the time (95% CI 85.2 to 91.1); on the disagreement stratum (n=1,651 adjudicated), the match rate was 60.4% (95% CI 58.0 to 62.7). The gap between the two strata is the evidence that routing by agreement concentrates human effort where it changes the label, and that a human on the agreement stratum still overturns roughly one label in nine rather than accepting all of them.

Across the validated set, 2,145 intents were confirmed and 255 spans were rejected as not prospective. The reject rate is itself an intent-level false-positive signal: it rises from 2.9% on the agreement stratum to 14.4% across the disagreement stratum (and 1.3% on the rare-modality stratum), so the spans the models disagree on are also the ones most likely not to be prospective actions at all. A second trained annotator independently relabeled a stratified 211-intent sample (rare axis values oversampled), blind to the first rater and to the models. Inter-annotator agreement is moderate to substantial: Cohen's kappa is 0.76 for type, 0.71 for modality (0.78 linearly weighted on the strength scale), 0.64 for action, and 0.58 for request-intent, and the two raters agreed on prospective inclusion for 96% of spans. The disagreements are the adjacency the models also show (start versus continue, plan versus order, ordered versus planned), not category confusion.

Human validation was performed at the span level over the four closed CIR axes (action, type, request-intent, modality) and target identity, which the annotator confirmed or corrected per field. A second validation pass then covered the six sparse attribute fields (time, frequency, duration, condition, reason, goal) on a 300-intent audit. On that 300-intent sample, the per-field human-validated accuracy (fraction correct among the values the annotator judged Yes/No) is reason 94%, duration 92%, frequency 86%, goal 84%, condition 77%, and time 45%. The time field is the clearest instance of the paper's central caution that agreement is not correctness: the three annotation models agree on timing about 90% of the time (Appendix A.1) yet are right only 45% of the time, because all three shared an over-generation bias, placing a value where the note states none, that consensus alone cannot catch. The human review corrected these and the released gold carries the corrected (usually blank) value. Because routing checks only the five critical fields, the sparse attribute fields receive no consensus protection, which is exactly why their accuracy is estimated by direct human audit rather than inferred from agreement.

### 7.2 Baseline model evaluation

We evaluate existing models against the human gold with no task-specific training, on the two benchmark tracks of §6.

**Span-level labeling (primary).** Every model receives the same 600 gold spans, a stratified sample of the 2,145 verified intents, each with its local context, and predicts the four closed CIR fields. There is no extraction step, so coverage is 100% and all models are compared on an identical set. We score each field, and the exact match of all four, against the human gold (Table 8).

| Model | action | type | request-intent | modality | all-four exact |
|---|---|---|---|---|---|
| Claude Sonnet 4.5 | 63 | 91 | 75 | 67 | 35 |
| Llama 3.3 70B | 48 | 85 | 72 | 59 | 25 |
| gpt-oss-120b | 46 | 90 | 70 | 60 | 20 |
| Nova Pro | 46 | 87 | 76 | 50 | 18 |
| gpt-oss-20b | 45 | 86 | 70 | 58 | 19 |

***Table 8.*** *Span-level labeling accuracy (%) against the human gold. All five models receive the same 600 spans with local context and predict the four closed fields, so the comparison is at full coverage and free of selection bias. Type is recovered reliably (85 to 91%); the two new axes are moderate (request-intent 70 to 76%, modality 50 to 67%); action is the hardest field (45 to 63%), dragged by the start-versus-continue distinction, which a local span cannot resolve because it requires the admission medication list (Appendix A.4). The all-four-exact score (18 to 35%) is therefore bounded by action. Sonnet leads on the all-four-exact score and on every field except request-intent, where Nova Pro is marginally higher (76 vs 75). Sonnet, Llama, and Nova also served in construction (§4), so they are construction-involved references; the two gpt-oss models are the independent baselines (all-four 19 to 20%), so the difficulty holds for models uninvolved in construction.*

**Whole-note note-only extraction (secondary).** This track measures recovering the spans *without* anchors: the model reads the note and must find the prospective spans itself. With the tuned benchmark prompt, models recover 73% to 87% of the anchor-derived gold note-only (Table 9). The benchmark prompt instructs the model to extract one intent per listed discharge medication; models do not enumerate the medication list unprompted, so this instruction is what carries medication recall. This recall is at an uncontrolled, lower-bounded precision (the strongest model extracted 3.3 times as many intents as the sampled gold), and it is measured against anchor-derived gold, so it does not by itself show that anchors are unnecessary. Anchoring remains an annotation-efficiency and coverage strategy. The results are consistent with three benefits of anchoring: localization, residual recall (the strongest model reaches 87% note-only, the others 73 to 74%), and completeness on the fine attributes the note-only setting systematically drops.

Table 9 reports the whole-note results across models. Type is recovered well by every model (94 to 97%), while request-intent and modality are the hardest fields for all models, and the open-weight gpt-oss-120b is competitive with the frontier models on this task. The matcher weights type at 0.2, so we re-scored this track with type removed (span/target only): type accuracy falls only about three points (Sonnet 97 to 94) and the model ranking is unchanged, so the matching selection does not materially inflate the type result.

| Model | recall | precision (lower bound) | action | type | request-intent | modality |
|---|---|---|---|---|---|---|
| Claude Sonnet 4.5 | 87 | 27 | 85 | 97 | 75 | 76 |
| Llama 3.3 70B | 74 | 26 | 77 | 96 | 77 | 72 |
| gpt-oss-120b | 74 | 29 | 81 | 97 | 76 | 75 |
| Nova Pro | 73 | 27 | 80 | 94 | 77 | 72 |
| gpt-oss-20b | 73 | 27 | 78 | 95 | 76 | 68 |

***Table 9.*** *Whole-note, note-only results (%) on the held-out test split (450 notes, 1,939 gold intents), using the tuned benchmark prompt and the §6 matching rule, with the tuning dev notes held out. Recall is the fraction of gold intents the model extracted; the field figures (action, type, request-intent, modality) are accuracy on matched intents. Precision is a lower bound: the gold is sampled per MIMIC note rather than exhaustively, so many extra extractions are likely unlabeled true intents rather than false positives (Sonnet extracted 6,331 intents against 1,939 sampled gold), which is why precision is reported only as a lower bound. A clean precision figure requires exhaustive per-note gold; §7.6 provides it on a 24-note subset (Table 14). Sonnet, Llama, and Nova also served in construction (§4); the gpt-oss models are the independent baselines and were run at low reasoning effort for throughput, so their figures are slightly conservative. Field accuracies here are conditional on the intents each model recovered and matched, so they are not directly comparable to the given-span accuracies of Table 8 (which score every supplied span).*

Beyond the coarse fields, the whole-note (note-only) setting reveals a systematic under-generation of the fine attributes. Scored against the human-corrected gold, models omit stated attributes far more often than they invent them: models state the condition trigger correctly on about 66% of matched intents and omit it on most of the rest; timing, frequency, and goal are dropped in the same way. With the benchmark prompt's instruction to extract every stated attribute, timing and frequency misses shrink, but condition stays under-extracted without an anchor. What the note-only setting shows here is that fine-attribute completeness, especially conditions, is where the anchor-free setting loses the most.

### 7.3 Error analysis

We break the span-level track (the 600-span primary set) down by corpus, by field value, and by the specific label confused, using the leading model (Sonnet 4.5) so the pattern is not obscured by a weaker model's noise. Each of the three findings below points at a demand the label makes on context rather than a modeling failure.

**Difficulty is corpus-dependent, and type is easy everywhere.** All-four-exact accuracy ranges from 29% on the assessment-and-plan corpus (ap_parsing) and the public ACI-Bench subset (SIMORD) to 55% on the patient-instruction corpus (PaniniQA), yet type is recovered at 86 to 100% in every corpus (Table 10). The losses concentrate where language is most hedged: ap_parsing, free-text assessment-and-plan prose, falls furthest on request-intent (62%) and modality (57%) because an A&P line states the plan without the explicit ordering verbs that fix authority and strength. The structured discharge corpora (MedDec, PaniniQA) carry those cues on the surface and score higher on both axes.

| Corpus | n | action | type | request-intent | modality | all-four exact |
|---|---|---|---|---|---|---|
| PaniniQA (patient instructions) | 38 | 68 | 100 | 92 | 82 | 55 |
| MedDec (discharge decisions) | 33 | 67 | 91 | 79 | 76 | 48 |
| CLIP (discharge follow-up) | 363 | 60 | 91 | 79 | 69 | 35 |
| ap_parsing (assessment & plan) | 145 | 68 | 89 | 62 | 57 | 29 |
| ACI-Bench / SIMORD (public) | 21 | 71 | 86 | 67 | 62 | 29 |

***Table 10.*** *Span-level per-field accuracy (%) for Sonnet 4.5 broken down by source corpus, on the 600-span primary set. Type transfers across corpora; the axes that encode authority (request-intent) and strength (modality) degrade most on the free-text assessment-and-plan corpus, whose prose omits the explicit ordering cues that structured discharge notes carry. Small per-corpus n (MedDec 33, PaniniQA 38, ACI-Bench 21) makes those cells indicative rather than precise.*

**Action errors are concentrated in the medication verbs and in one reconciliation-only distinction.** Action accuracy is high for verbs with surface cues (stop 100%, refer 92%) and low for verbs that need document- or list-level context (obtain 25%, adjust 35%, perform 50%, instruct 55%). The largest single error is the start-versus-continue flip: continue is labeled start 49 times and start is labeled continue 21 times, together about a third of Sonnet's roughly 222 action errors. The pipeline resolves this distinction with the deterministic medication-reconciliation layer (Appendix A.4), which compares the admission and discharge lists; a local span carries no such list, so the model cannot in principle recover it. The next-largest flips, follow_up labeled schedule (25) and obtain labeled perform (10), are order-versus-execute ambiguities the note text leaves underspecified.

**Modality errors are adjacent-strength slips, not category errors.** Every frequent modality confusion sits between neighboring points on the strength scale: ordered labeled planned (39) or recommended (28), planned labeled recommended (21), optional labeled conditional (20). The model rarely crosses from an active strength to a prohibition; it mis-grades the degree of commitment, exactly the judgement the modality axis was introduced to capture and the one hedged prose makes hardest. The gold is dominated by majority classes (order 393 of 600 for request-intent, continue 212 and medication 325 for action and type), so all-four-exact is driven mainly by the common intents behind downstream care actions, and the low headline reflects the action field (the start-versus-continue bound above) far more than a broad failure across the schema.

**Setting aside the reconciliation-only action distinction recovers most of the gap.** Because start, adjust, and continue cannot be told apart from a local span (they require the admission medication list, §4), we re-score the 600 spans with those three verbs collapsed into a single medication-management class, keeping stop and every other verb distinct (Table 11). Action accuracy rises from 45 to 63% to 72 to 79%, and all-four-exact from 18 to 35% to 38 to 47%, across all five models. The exact verb is retained in the release and remains the primary score (Appendix A.5); the collapse is an action-granularity sensitivity analysis.

| Model | action exact | action (superclass) | all-four exact | all-four (superclass) |
|---|---|---|---|---|
| Claude Sonnet 4.5 | 63 | 78 | 35 | 47 |
| gpt-oss-120b | 46 | 79 | 20 | 46 |
| Nova Pro | 46 | 76 | 18 | 38 |
| Llama 3.3 70B | 48 | 72 | 25 | 44 |
| gpt-oss-20b | 45 | 76 | 19 | 43 |

***Table 11.*** *Span-level action and all-four accuracy (%) with start / adjust / continue collapsed into one medication-management class (stop and all other verbs kept distinct), computed on the same 600 predictions as Table 8, so the exact-action and all-four columns are identical to it. Collapsing the one distinction a local span cannot resolve recovers 15 to 33 points of action accuracy, confirming that action, not the schema as a whole, bounds the all-four score.*

### 7.4 Results on target and sparse attributes

The four closed fields are not the whole record: target is required, and timing and condition carry the paper's clinical-tracking value. We score them on the same 600 spans (Table 12). Target-category (the normalized concept) is recovered on 40 to 62% of spans, the hardest required field, because a span names a drug or test in many surface forms. Standard target codes cannot be scored here: the validated gold does not yet carry resolved LOINC/SNOMED/RxNorm codes, so code-level benchmarking waits on a coded gold pass. For the sparse fields the difficulty decomposes cleanly: models get the *value* right when a value is truly present (time 75 to 84%, condition 88 to 96% on true-present cases), which suggests the low aggregate time accuracy of §7.1 is driven by *presence* (over- and under-generation; time presence F1 49 to 60, condition 71 to 77) rather than by normalization: conditional on detecting the attribute, the value is usually right. This is a more actionable target for future models than a single blended accuracy.

| Model | target-category | time presence F1 | time value | condition presence F1 | condition value |
|---|---|---|---|---|---|
| Claude Sonnet 4.5 | 54 | 55 | 84 | 77 | 96 |
| Llama 3.3 70B | 46 | 52 | 76 | 74 | 92 |
| gpt-oss-120b | 62 | 60 | 81 | 71 | 93 |
| Nova Pro | 40 | 49 | 80 | 71 | 94 |
| gpt-oss-20b | 52 | 52 | 75 | 74 | 88 |

***Table 12.*** *Target and sparse-attribute accuracy (%) on the 600-span set. Target identity is human-validated for all spans; the time and condition gold are the released annotation values, human-audited on a 300-intent subset (§7.1). Presence F1 is precision/recall of stating the attribute at all; value accuracy is on cases where both gold and model state a value. Timing and condition values are recovered accurately when present; the bottleneck is deciding whether the span carries them, which is where models over- and under-generate.*

### 7.5 Cross-corpus transfer baseline

Harmonization makes a transfer test possible that no single source corpus can pose: train on four corpora and test on the fifth. We train a light TF-IDF plus logistic-regression classifier on the human-gold spans of four corpora and test on the held-out fifth, for type, request-intent, and modality, alongside a pooled in-distribution five-fold reference (Table 13). The classifier uses word 1-2-gram TF-IDF (min document frequency 2) and balanced-class logistic regression on the span text, with no per-corpus tuning; the in-distribution folds are pooled across corpora. Every field shows a transfer gap: in-distribution to leave-one-corpus-out drops 79 to 71% (type), 75 to 68% (request-intent), and 70 to 61% (modality). The gap is not uniform: the held-out assessment-and-plan corpus (ap_parsing) and the held-out dialogue corpus (SIMORD) fall furthest (type 43% and 54%), the same two distributions the error analysis flagged as least explicit. This simple lexical baseline therefore shows a measurable cross-corpus generalization gap; how much of it reflects language, label-prior, or annotation-distribution shift needs stronger models and class-balanced metrics to separate.

| Field | in-distribution (5-fold) | held-out mean | CLIP | MedDec | ap_parsing | PaniniQA | SIMORD |
|---|---|---|---|---|---|---|---|
| type | 79 | 71 | 80 | 92 | 43 | 86 | 54 |
| request-intent | 75 | 68 | 76 | 85 | 50 | 80 | 49 |
| modality | 70 | 61 | 68 | 75 | 45 | 76 | 42 |

***Table 13.*** *Cross-corpus transfer accuracy (%) of a TF-IDF plus logistic-regression classifier trained on the human-gold spans of four corpora and tested on the held-out fifth, versus a pooled in-distribution five-fold reference. The 7 to 9 point in-distribution-to-held-out accuracy gap, concentrated on the free-text assessment-and-plan and dialogue corpora, quantifies the cross-corpus generalization gap a harmonized resource makes measurable. Held-out macro-F1 (mean across corpora) is lower, at 42 (type), 44 (request-intent), and 52 (modality), reflecting the heavy class imbalance, so part of the accuracy gap is a shift in class priors rather than pure linguistic transfer difficulty.*

### 7.6 Whole-note validation

To measure what the sampled, anchor-derived gold cannot, a trained annotator exhaustively annotated 24 whole notes from scratch across both distributions, blind to the source anchors and the model outputs (330 prospective intents). The released candidate pool contains 96.4% of these exhaustive intents, so only 3.6% are missed by both the source anchors and the three models. Scored against this exhaustive gold, the models reach the true precision the sampled test set could only lower-bound (Table 14): 78 to 96% precision at 61 to 80% recall. True field prevalence on these notes is 29% for timing and 21% for condition, close to the annotation fill-rates of Figure 2, which supports that the sparse fields are genuinely sparse rather than only under-annotated.

| Model | recall | precision | F1 |
|---|---|---|---|
| Claude Sonnet 4.5 | 80 | 87 | 83 |
| Llama 3.3 70B | 77 | 85 | 81 |
| gpt-oss-120b | 70 | 89 | 78 |
| Nova Pro | 61 | 78 | 69 |
| gpt-oss-20b | 75 | 96 | 84 |

***Table 14.*** *Whole-note extraction against exhaustive from-scratch human gold on 24 notes (330 intents), stratified across both distributions. Unlike the sampled test set (Table 9), this gives true precision rather than a lower bound. Recall is the fraction of exhaustive human intents the model recovered; precision the fraction of its extractions that hit one. Small n (24 notes) makes these indicative.*

## 8 Access, provenance, and licensing

### 8.1 Intended uses

CIRCA supports three things the source corpora cannot do individually: extract prospective actions with their authority and clinical strength under one label space; transfer across annotation schemes and note types; and test whether models preserve clinically consequential distinctions such as conditional-versus-active and prohibited-versus-recommended. The provenance tiers add a fourth: training on the full 10,011-intent resource while evaluating on the human-reviewed core, or studying robustness to silver-label noise. Three concrete uses motivate the release. **(a) Lost-to-follow-up safety:** annotations with a validated timing value become trackable deadlines, and the modality axis lets a tracker suppress reminders for prohibited or not-yet-triggered conditional intents; an untracked "repeat CT in 3 months if the nodule persists" is exactly the failure mode behind missed-follow-up harm. **(b) FHIR request-resource serialization:** request-intent maps to FHIR RequestIntent and the coded target to the resource, so a CIR record fills in a ServiceRequest, MedicationRequest, or CommunicationRequest entry. **(c) Cross-corpus robustness:** the five-corpus, two-distribution design supports training on discharge summaries and testing on dialogue-derived notes, an evaluation only a harmonized resource enables.

CIR output is a structured extraction, not a clinical order, and schema validity is not clinical correctness: an erroneous extraction of a prohibited or conditional action could cause harm if consumed unchecked. The resource is intended for research and system development, not autonomous clinical use.

### 8.2 Data provenance and compliance

Every intent in CIRCA traces to its source note and to the policy under which that note was handled.

**Sources.** Four corpora (CLIP, MedDec, ap_parsing, PaniniQA) annotate MIMIC-III notes, protected clinical records available only under credentialed PhysioNet access and the MIMIC-III data-use agreement (DUA). The fifth, ACI-Bench, is public, simulated doctor-patient dialogue and contains no real patient data.

**Processing.** The notes were never sent to an uncontrolled model endpoint: every annotation-model call ran through Amazon Web Services (AWS) Bedrock under a HIPAA Business Associate Agreement (BAA), on a zero-retention configuration under which no prompt or model output is stored once a request completes. MIMIC-III is de-identified, so a BAA is not strictly required for it; we adopt one as an additional contractual safeguard. This covers the processing infrastructure only and is applied on top of, not in place of, each source corpus's own DUA.

**De-identification handling (KART).** MIMIC notes carry de-identification placeholders (for example [**Last Name**] or [**2118-6-2**]) where identifiers were removed. Passing these to a model degrades reading and destroys the time structure that timing extraction depends on, so before model processing we fill each placeholder with a realistic, deterministic surrogate using KART, a surrogate filler extended for discharge summaries and released with the code. KART classifies every placeholder against the PhysioNet de-identification tag grammar and substitutes a synthetic value under two invariants: *consistency* (a given placeholder maps to the same surrogate throughout a note) and *interval preservation* (all full dates in a note shift by one shared year offset, keeping month and day, so the gaps between dated events are preserved for timing extraction). The released version handles the failure modes a discharge corpus exposes, including date-label echoing, numeric-identifier misrouting, a dedicated initials surrogate, and distinct surrogate names for id-less clinician mentions while protecting the patient's own name; across 100 notes and 4,632 placeholders none was left unclassified. Surrogate values are synthetic and carry no real identifiers; the underlying clinical text remains MIMIC-derived and is processed only under the BAA path above.

**Release.** We publish only derived artifacts, never source note text: the CIR labels and annotations, the source-to-CIR crosswalks, and the per-note KART surrogate mappings. The MIMIC-derived layers ship as stand-off annotations, character offsets and CIR fields keyed to MIMIC note identifiers, together with the surrogate mapping (placeholder offsets and their synthetic fills) and a rebuild script that regenerates the exact text the models read for a user who already holds credentialed MIMIC access (Table 15). Only the ACI-Bench layer, built on public text, is released with its note text.

### 8.3 Availability and licensing

The deposit ships as stand-off CIR records in JSONL, one object per intent with a note identifier, character-offset provenance, and all CIR fields, together with the crosswalks, the KART surrogate mappings, the deterministic FHIR mapper, and the reference evaluator. Each layer inherits the license and data-use agreement of its source corpus (Table 15): the combined resource does not carry a single uniform license. The complete deposit is archived on Zenodo at https://doi.org/10.5281/zenodo.22058593.

| Layer | underlying-text rights | annotation license | redistributable? | access mechanism |
|---|---|---|---|---|
| CLIP | MIMIC-III (PhysioNet DUA) | source corpus license | annotations only, no MIMIC text | PhysioNet credentialed + rebuild script |
| MedDec | MIMIC-III (PhysioNet DUA) | source corpus license | annotations only, no MIMIC text | PhysioNet credentialed + rebuild script |
| ap_parsing | MIMIC-III (PhysioNet DUA) | source corpus license | annotations only, no MIMIC text | PhysioNet credentialed + rebuild script |
| PaniniQA | MIMIC-III (PhysioNet DUA) | source corpus license | annotations only, no MIMIC text | PhysioNet credentialed + rebuild script |
| SIMORD / ACI-Bench | ACI-Bench (public, simulated, CC BY 4.0) | CDLA-Permissive-2.0 | text and annotations | direct download |
| CIR annotation layer (ours) | per source layer above | CC BY 4.0 | annotations + crosswalks + code | repository |

***Table 15.*** *Per-layer rights and access. The combined resource does not carry a single uniform license: each layer inherits the rights of its underlying text, and only the SIMORD/ACI-Bench layer can be redistributed with its text. The mapper, evaluator, and crosswalks are released under an open code license.*

## 9 Limitations

Construction relies on model agreement, and the human audit (§7.1) calibrates that agreement against correctness. For the five critical fields, which routing checks on every intent, audited accuracy is 88.4%. For the sparse attribute fields, which routing does not check, agreement and correctness can diverge: the time field shows about 90% inter-model agreement against 45% human-validated accuracy (§7.1), because all three models shared an over-generation bias. This divergence is why the release pairs agreement statistics with the audit, and why the audit, not agreement, is the accuracy estimate to cite for sparse fields.

The construction measurements have defined scope. Inter-model agreement is computed on 8 to 40 documents per corpus (Appendix A.1), with correspondingly wide intervals except in the scaled temperature-0 run (Appendix A.3); extending the scaled run to the remaining corpora is a direct next step. The medication arbiter applies to notes carrying structured admission and discharge medication lists, which the assessment-and-plan corpus does not include. Source annotations serve two roles, as evidence shown to annotators and as recall targets, so a no-source-hint ablation is the designated experiment for separating genuine recall from hint propagation.

The main benchmark evaluates prompted models against the human gold (§7.2). Sonnet, Llama, and Nova contributed candidate labels during construction, so they are construction-involved references; the two gpt-oss models are the independent baselines (§4). A TF-IDF plus logistic-regression cross-corpus baseline (§7.5) establishes a clear transfer gap; a strong trainable-encoder baseline on the released split is future work.

The gold standard is structured by field. All 2,145 intents are human-checked on the five critical fields; the six sparse fields are human-checked on a 300-intent audit, and target codes are not yet resolved in the gold (§7.4). Records are therefore core-field gold, with sparse-field and code quality quantified by the audits. Per-corpus notes, patients, intents, gold, and routing are reported in Table 6, with split sizes in §6; a full dataset card will add candidate-versus-clustered intent counts. On three corpora the retrospective, over-segmented source gold bounds measurable recovery of some source labels, a property we characterize in Appendix A.2.

Span coverage has a measured ceiling. Spans come from the source anchors plus LLM extraction, so an intent missed by both the source corpora and the models goes uncounted; the exhaustive from-scratch pass of §7.6 estimates this miss rate at 3.6% on 24 notes (completeness 96.4%), and a larger exhaustive study will tighten the estimate.

On standards and human validation, the released mapper emits schema-valid FHIR R4 resources for every mappable intent (§3.2); profile-conformance and terminology-binding validation are future work. Reliability is measured by a second annotator relabeling a 211-intent sample (§7.1, moderate-to-substantial kappa); full double annotation of the release and validation by expert clinicians are the next stages of the validation program.

## 10 Conclusion

Prospective clinical intent is annotated in fragments across incompatible corpora. We defined the CIE task and a FHIR-informed representation that adds request-intent and modality as separate axes prior datasets do not jointly represent, and we release CIRCA, a harmonized dataset of 10,011 intents over 1,185 notes and 942 patients across five heterogeneous corpora and two note distributions, with a human-reviewed core of 2,145 validated intents plus 255 reviewed negatives, per-layer licensing, source-to-CIR crosswalks, and a schema-valid FHIR R4 mapper. The dataset is built by a three-model consensus that routes disagreements to human adjudication; the audited agreement stratum matches human judgment 88.4% of the time. Benchmarking five existing models against the human gold shows type is easy (85 to 91%) but the full structured label is not (18 to 35% all-four given the span), with action and modality the hardest fields. All artifacts are released; an exhaustive from-scratch whole-note gold set (§7.6) and a second-annotator reliability study (§7.1) are included; validation by expert clinicians remains future work.

## Data availability

CIRCA is archived on Zenodo at https://doi.org/10.5281/zenodo.22058593. The deposit contains the CIR annotations, the source-to-CIR crosswalks, the per-note KART surrogate mappings, the modified KART code, the deterministic FHIR R4 mapper, the reference evaluator, and a rebuild script. It contains no source note text: the MIMIC-III-derived layers ship as stand-off annotations (character offsets and CIR fields keyed to MIMIC note identifiers) that a credentialed user rehydrates with the rebuild script and their own MIMIC-III access, and only the public ACI-Bench layer is distributed with its text. Use of the MIMIC-III-derived layers remains governed by the PhysioNet/MIMIC-III data-use agreement.

## Appendix A. Construction and validation details

### A.1 Inter-model agreement

Table 16 reports how often the three annotation models agree on each field, per corpus, on aligned clusters. Agreement on target, type, and the two axes holds across both MIMIC discharge summaries and ACI-Bench dialogue-derived notes; action is reported as exact and polarity-preserving agreement (Appendix A.5), the latter crediting a start-versus-continue disagreement that preserves the same direction of care. Inter-model agreement measures consistency of the construction pipeline, not correctness; the human validation of §7.1 is the measure that certifies the release.

| Corpus (note type) | target | time | type | modality | req_intent | action* |
|---|---|---|---|---|---|---|
| CLIP (MIMIC discharge) | 95 | 95 | 97 | 90 | 92 | 85 · 99 |
| MedDec (MIMIC discharge) | 96 | 88 | 88 | 92 | 100 | 85 · 92 |
| ap_parsing (MIMIC assessment-and-plan) | 98 | 92 | 87 | 94 | 96 | 76 · 91 |
| PaniniQA (MIMIC instructions) | 89 | 97 | 93 | 87 | 98 | 84 · 90 |
| SIMORD (ACI-Bench dialogue) | 82 | 92 | 89 | 95 | 84 | 84 · 92 |

***Table 16.*** *Inter-model field agreement (%) on aligned clusters; representative runs of 8 to 40 documents per corpus. The action column reports two numbers: exact · polarity-preserving. Agreement on the core fields holds across both MIMIC discharge summaries and ACI-Bench dialogue-derived notes. A scaled temperature-0 run (Appendix A.3) narrows the intervals.*

### A.2 Recovery of source gold

We measure recall of each corpus's own annotations by the harmonized output and the accuracy of the adapted type (Table 17). Recall is the metric this partial gold supports: each corpus annotates only a slice of the intents in a note, so intents the pipeline adds beyond the gold are the point of harmonization, not false positives. Because source annotations are also shown to the annotators as evidence (§4), this recall measures consistency with the input signal; independent validation comes from the human gold study (§7.1); a no-source-hint ablation (§9) will separate genuine recall from hint propagation. The lower recall on three corpora reflects source-gold granularity rather than misses: of MedDec's unmatched source labels, 57% (174 of 306) are Category 3 (Defining problem) diagnoses out of CIE scope and 25% carry retrospective markers (s/p = status post, h/o = history of, history, received); CLIP's unmatched labels are dominated by appointment logistics and section headers; and PaniniQA includes past care. Source-gold recall is therefore a lower bound on prospective-action recall, not a miss rate.

| Corpus | gold granularity | recall | type acc. |
|---|---|---|---|
| ap_parsing | discrete action items | 83 | 87 |
| SIMORD | discrete orders | 77 | 94 |
| CLIP | sentence-level (over-segmented) | 42* | 86 |
| PaniniQA | span-level (incl. past care) | 45* | 62 |
| MedDec | decision spans (retrospective) | 9* | 71 |

***Table 17.*** *Recall of each corpus's own gold labels (%). On the two corpora whose gold is discrete and prospective (ap_parsing, SIMORD), recall is 77 to 83% with 87 to 94% type accuracy. *Lower figures on CLIP, PaniniQA, and MedDec reflect source-gold granularity (over-segmented, retrospective, or past-care spans); source-gold recall is a lower bound on prospective-action recall, not a miss rate.*

### A.3 Decoding and confidence intervals

Small annotation sets are noisy: one hard document can move an aggregate by ten points. We therefore report inter-model agreement with a document-level bootstrap confidence interval. Brackets in this subsection denote 95% bootstrap confidence intervals. At temperature 0 on a scaled run (ap_parsing, 20 documents, 169 clusters), agreement carries tight intervals: target 98% [96, 100], type 87% [82, 92], modality 94% [89, 98], and polarity-preserving action 88% [83, 92]. The same temperature-0 run gives an exact action agreement of 69% [60, 77]; exact agreement falls below the polarity-preserving rate because it penalizes the start-versus-continue flips that preserve the direction of care. We report exact action agreement as the primary action figure and the polarity-preserving rate as a secondary, clinical-severity-aware measure alongside it.

### A.4 Deterministic normalization layers

Deterministic layers convert model-specific surface variation into a canonical form before comparison, so that agreement reflects genuine disagreement rather than formatting:

- **Temporal.** The model emits the verbatim time phrase; a script performs all normalization to a day range ("in three weeks", "P21D", and "P3W" collapse to the same value), removing the model's own ISO inconsistency.
- **Polarity.** Unconditional negative directives ("avoid", "do not", "no lifting") map modality to *prohibited*, so a flat prohibition is not scored as a recommendation. A conditional hold instruction ("hold for INR above 3.5") is treated differently: it preserves the trigger as a stop-condition and keeps its active modality rather than becoming prohibited.
- **Medication reconciliation.** Parsing the note's admission and discharge medication lists yields a reconciliation-derived action label (start / adjust / continue / stop). It is applied only as an *all-or-nothing cluster arbiter*: the reconciliation verb overrides a cluster's action only when every model's extraction points at the same drug and the lists give one answer, so it can settle a disagreement but never manufacture one (Appendix A.5).
- **Terminology coding.** Codeable targets resolve to LOINC / SNOMED / RxNorm via public services; codes drive a cannot-link constraint in alignment (two intents with different resolved drug codes never merge).

- **Canonicalization.** Closed action vocabulary with synonym mapping, brand-to-generic target normalization, and coarse strength classes for the fine modality values that discharge language genuinely conflates.

### A.5 Polarity-preserving action metric and component attribution

The action verb is where models legitimately differ, so scoring benign adjacency (start vs adjust) like real error (start vs stop) understates agreement. We freeze a clinical distance matrix in advance (medication-management verbs near each other, any verb versus *stop* maximally distant) and report the polarity-preserving rate, agreement within that threshold, as a secondary measure beside the primary exact rate.

Component attribution uses leave-one-out paired-delta testing: a component counts as an improvement only when the paired confidence interval of its effect excludes zero. The code-aware cannot-link constraint splits clusters whose members carry conflicting standard codes, correcting spuriously high agreement. The medication arbiter unifies the action label across models, so we validate it against human gold: on the 1,221 medication intents in the validated set, action accuracy is 85% both before and after, because start versus continue is decided by the admission medication list the arbiter reads but the span-level annotator did not see. Its contribution is cross-model consistency at unchanged accuracy.